\documentclass[conference]{IEEEtran}
\IEEEoverridecommandlockouts
\usepackage{cite}
\usepackage{amsmath,amssymb,amsfonts}
\usepackage{algorithmic}
\usepackage{graphicx}
\usepackage{textcomp}
\usepackage{xcolor}
\usepackage{hyperref}
\def\BibTeX{{\rm B\kern-.05em{\sc i\kern-.025em b}\kern-.08em
    T\kern-.1667em\lower.7ex\hbox{E}\kern-.125emX}}
\begin{document}

\begin{titlepage}
\centering
\vspace*{\stretch{1}} 
IEEE Copyright \copyright\ 2023. Personal use of this material is permitted.\\
 Permission from IEEE must be obtained for all
other uses, in any current or future media, including reprinting/republishing this material for advertising
or promotional purposes, creating new collective works, for resale or redistribution to servers or lists, or
reuse of any copyrighted component of this work in other works.
\vspace*{\stretch{1}} 

Accepted to be Published in: Proceedings of the 2024 IEEE PES General Meeting,\\
21-25 July 2024, Seattle, WA, USA.
\vspace*{\stretch{2}}
\end{titlepage}

\title{Transformer meets wcDTW to improve real-time battery bids: A new approach to scenario selection
}


\author{\IEEEauthorblockN{Sujal Bhavsar}
\IEEEauthorblockA{\textit{Ascend Analytics} \\
Boulder, USA \\
\href{https://orcid.org/0000-0002-9845-2106}{0000-0002-9845-2106}}
\and
\IEEEauthorblockN{Vera Zaychik Moffitt}
\IEEEauthorblockA{\textit{Ascend Analytics} \\
Boulder, USA \\
vmoffitt@ascendanalytics.com}
\and
\IEEEauthorblockN{Justin Appleby}
\IEEEauthorblockA{\textit{Ascend Analytics} \\
Boulder, USA \\
jappleby@ascendanalytics.com}
}

\maketitle

\begin{abstract}
Stochastic battery bidding in real-time energy markets is a nuanced process, with its efficacy depending on the accuracy of forecasts and the representative scenarios chosen for optimization. In this paper, we introduce a pioneering methodology that amalgamates Transformer-based forecasting with weighted constrained Dynamic Time Warping (wcDTW) to refine scenario selection. Our approach harnesses the predictive capabilities of Transformers to foresee Energy prices, while wcDTW ensures the selection of pertinent historical scenarios by maintaining the coherence between multiple uncertain products. Through extensive simulations in the PJM market for July 2023, our method exhibited a 10\% increase in revenue compared to the conventional method, highlighting its potential to revolutionize battery bidding strategies in real-time markets.
\end{abstract}


\section{Introduction}
The energy landscape is witnessing a paradigm shift driven by the increasing integration of renewable sources, such as wind and solar. As these intermittent energy sources become more predominant, the emphasis on grid-level storage to counteract supply fluctuations is growing exponentially. Such systems play a crucial role in maintaining grid stability and managing the variability of renewable sources. Deregulated global energy markets are undergoing transformative changes with the surge in grid-level storage adoption \cite{schleicher2012renewables}. Storage mechanisms not only stabilize the grid but also provide a unique financial opportunity through energy arbitrage. This process, leveraging the buy-store-sell mechanism in response to electricity price variations, is emerging as a vital component in justifying investments in grid-level storage. The efficacy of the investment hinges on the precision of the battery's bidding strategies \cite{krishnamurthy2017energy,herding2023stochastic}.

The intricate nature of bidding in the electricity market stems from the inherent challenge of maintaining a delicate equilibrium between supply and demand. To adeptly navigate this complexity, Independent System Operators (ISOs) and Regional Transmission Organizations (RTOs) employ multi-settlement market structures \cite{FERC2023, bhattacharyya2019energy}. These structures encompass a spectrum of time horizons, from day-ahead predictions to real-time adjustments. It is noteworthy that real-time markets, in their bid to reflect immediate dynamics, settle based on intra-hour pricing, commonly segmented into 5, 10, or 15-minute intervals. These settlements occur based on the bids provided by the participants. Certain markets, such as NYISO, PJM, CAISO, and ERCOT, to name a few, not only make settlements in real-time based on previously submitted bids but also facilitate real-time bid submission. In these setups, market participants have the latitude to submit bids every hour, targeting delivery in the following hour, or in the case of ERCOT, the current hour \cite{CAISO, ERCOT, PJM}.

The challenge of formulating optimal bids that maximizes revenue by avoiding penalties in real-time electricity markets, especially in the face of uncertain market dynamics and fluctuating renewable generation, can be adeptly addressed through stochastic optimization. This framework seamlessly integrates probabilities and uncertainties associated with event occurrences. Such innovative theoretical advancements serve as the cornerstone for cutting-edge methodologies, and they have catalyzed a surge of research and literature in this domain \cite{zakaria2020uncertainty, bhavsar2023hybrid, herding2023stochastic, silva2022multistage}.
A challenge inherent in this approach is pinpointing a subset of future scenarios that captures the plausible distribution, ensuring the optimization framework yields dependable decisions. The complexity amplifies when these scenarios span multiple periods and incorporate a myriad of interdependent uncertain variables.

    A substantial body of literature exists addressing the creation of uncertain scenarios in electricity markets, considering both single and multiple variables across various timeframes. Many approaches, grounded in heuristics, select historical days without accounting for foresight \cite{gomes2017stochastic, akbari2019stochastic}, thereby making an implicit assumption that future market dynamics will mirror historical patterns. While some studies opt for random selection mechanisms \cite{fazlalipour2019risk}, others employ deterministic metrics like Euclidean distance to refine their subset, risking misplaced confidence in potentially suboptimal selection \cite{laws2022stochastic}. Herding et al. \cite{herding2023stochastic} proposes the use of SARIMA model, which, despite its prowess in time series forecasting, remains contingent on accurate hyperparameter identification and may falter in capturing non-linear relationships or unexpected market shocks. A noteworthy subset of literature incorporates forecasting methods to generate synthetic scenarios using univariate normal distributions of error, a choice that might oversimplify real-world error dynamics. Several studies have enhanced the reliability of synthetic scenarios for a single uncertainty using sophisticated deep learning techniques\cite{chen2018model}. Though such approaches are deemed more effective in estimating the future than the previous analogue scenarios based methods, an over-reliance on these forecasts can be precarious, especially in contexts where preserving interdependencies among multi-uncertainty is paramount. Therefore, this study exclusively concentrates on improving the selection of analogous scenarios from historical data and offers a comparison with the state-of-the-art method in this specific area.

In summary, in scenario selection for optimizing battery bids in real-time markets, traditional methods have predominantly hinged upon heuristics for historical day selection or the superimposition of random errors on forecasts, casting doubts over their encompassing reliability. Such methods might inadvertently sidestep certain pivotal scenarios, potentially undermining the robustness of derived outcomes. However, the advent of Transformer-based time series forecasting has ushered in a new era of enhanced prediction accuracy \cite{zeng2023transformers}. It should be noted that the term ``Transformer" in this context refers to the deep learning model, not to a physical transformer. Simultaneously, the imperative to adeptly cherry-picking analogous days from historical datasets has showed promise. This manuscript introduces an innovative method that melds the prowess of transformer-based forecasting with the precision of the proposed Weighted Constrained Dynamic Time Warping (wcDTW) metric. The aim is to pinpoint those historical days that best mirror contemporary dynamics.

\section{Methodology}

\begin{figure}[htbp]
\centering
{\includegraphics[width=90mm]{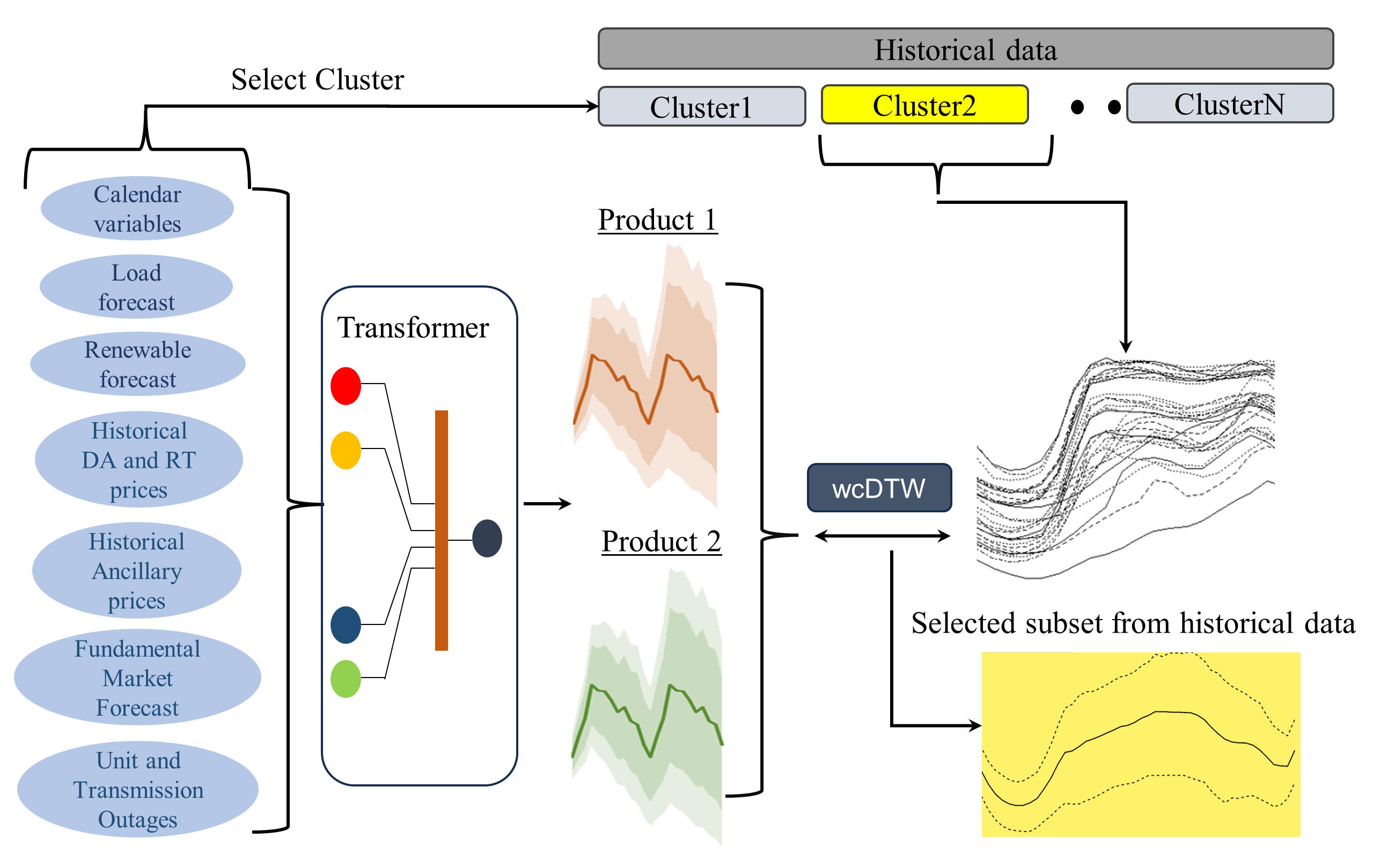}}
\caption{Schematic representation of the proposed approach.}
\label{fig1}
\end{figure}

In Figure \ref{fig1}, we present a schematic of the proposed methodology. At its core, this approach clusters  historically analogous days — both in terms of similarity and dissimilarity — employing refined clustering techniques as delineated by Bhavsar et al. \cite{bhavsar2021machine}, Panda et al. \cite{panda2021data}, and further expanded upon by \cite{elhamifar2015dissimilarity}. As depicted, the clustering mechanism ingests a spectrum of parameters related to market conditions. This includes calendar attributes, load and renewable forecasts, historical day-ahead (DA) and real-time (RT) prices, and prior ancillary price data. Furnished with these input features, and invoking diversity measures upon the target variables (which will be rendered as uncertain entities in the stochastic optimization process), this phase orchestrates the clustering of the historical dataset, ensuring it encapsulates the similarity patterns inherent in the data. A cluster corresponding to the given market condition is picked for further analysis, as illustrated by the highlight in Cluster 2 in Fig. \ref{fig1}.

Figure \ref{fig1} further elucidates how the market parameters serve as inputs to the Transformer-based model \cite{vaswani2017attention} to provide real-time forecasts. Historically, Transformer architectures have pioneered advancements in tasks within the Natural Language Processing (NLP) realm, particularly where capturing temporal dynamics is paramount. This trait makes such models uniquely suited for forecasting tasks, especially in domains like electricity pricing, where intricate temporal relationships exist. In the context of stochastic optimization for electricity pricing, the Transformer's ability to discern subtle yet crucial, time-bound correlations and dependencies offers unparalleled advantages. Notably, our design renders the output of the Transformer model probabilistic in nature. Figure \ref{fig1} highlights scenarios where decisions pertain to two uncertain products for battery bidding. The forecasts for these products are illustrated through a median trajectory, flanked by a lightly shaded regions, denoting a specific quantile range, thus providing a comprehensive probabilistic perspective.

Employing probabilistic forecasts for uncertain products is paramount when identifying a representative subset from a selected cluster. Notably, clusters can encompass hundreds or even thousands of timestamps, rendering them impractical for real-time stochastic optimization. To effectively reduce the cluster size while honoring the probabilities associated with extreme events and likelihood measures, it is crucial to not solely depend on data-driven cluster reduction techniques, as suggested by Bhavsar et al. \cite{bhavsar2021machine}, or model-based strategies like those presented by Dupavcova et al. \cite{dupavcova2003scenario}. Instead, our proposed method innovatively incorporates real-time forecasts alongside the wcDTW measure, spanning various quantiles and products. Selection from the cluster hinges on a newly introduced distance coefficient, rooted in the wcDTW metric. Entries with the lowest $D_c$ are selected, their probability being inversely proportional to this metric. The formula for the proposed $D_c$ is:

\begin{equation}
    D_c = \sum_p wcDTW_p(n,m) \cdot wf_p
\end{equation}

Where $D_c$ is the distance coefficient for a single historical timestamp, $p$ is an index variable for various products, $wf_p$ is the weight factor corresponding to product p, which is either user-defined or can be optimized in backtesting. $n$ and $m$ represent the lengths of two time-series being compared, which correspond to the dimensions of the $wcDTW$ matrix. $wcDTW_p(n,m)$ is a distance measure between two time series of product $p$. $wcDTW$ matrix is populated as:

\begin{equation}
\begin{aligned}
wcDTW(i, j) = \begin{cases}
\infty & \text{if } |i - j| > W, \\
\begin{array}{@{}l@{}}
w(i, j) + \min \bigl\{ wcDTW(i-1, j), \\
\qquad wcDTW(i-1, j-1), \\
\qquad wcDTW(i, j-1) \bigr\}
\end{array} & \text{otherwise.}
\end{cases}
\end{aligned}
\end{equation}

where:

\begin{equation}
    w(i,j) = \sum_q w^q \cdot d(x_i,y_j^q)
\end{equation}
\begin{itemize}

\item $d(x_i,y_j^q)$  is euclidean distance between $x_i$ and $q$th quantile of $y_j$.
\item $x_i$ is the $i$th element of time-series from historical timestamp and $y_j^q$ is the $j$the element of quantile $q$ of forecast.
\item $W$ is the window size constraint.
\item $w^q$ is the weight of element $y_j^q$ based on their respective quantiles.

\end{itemize}

The weight assigned to each quantile can either be user-defined or optimized through backtesting. The proposed $wcDTW$ is based on the  $DTW$ concept, utilizing recursion to calculate the distance between time series. For more information on derivation, readers are directed to the DTW literature \cite{senin2008dynamic}.

\section{Results and Discussion}

The effectiveness of the proposed methodology is assessed within the context of intra-day battery bidding, wherein decision-makers aim to adjust the bids placed in the day-ahead market based on the latest information about their energy production. The stochastic framework employed for evaluation closely aligns with the one delineated in \cite{silva2022multistage} for optimizing intra-day bidding. For the sake of simplicity and without loss of generality, both \(w_{f_p}\) and \(w^q\) are user-specified in this analysis. Specifically, \(w_{f_p}\) is defined as a one-hot encoded vector with emphasis solely on the Energy RT product. Conversely, \(w^q\) is assumed to follow a normal distribution, with a higher concentration around medium values and diminishing weights assigned to the extremes.

In evaluating the efficacy of our proposed components, we establish a benchmark that selects historical days based on the similarity of market conditions alone, without incorporating the Transformer and wcDTW layers.
This benchmarking strategy is consistent with leading methodologies for analogue scenario selection in literature, as demonstrated by Gomes et al. \cite{gomes2017stochastic} and Bhavsar et al. \cite{bhavsar2024stochastic}. For a robust assessment of the scenario set against actual values, we introduce a preliminary evaluation metric—termed the 'stoch metric' (\(SM\)). This metric captures both the mean deviation of scenarios from true values and their capability to encompass extreme values."

\begin{align}
\mu_{\text{error}} &= \left| T - \frac{\sum_{i} W_i \times F_i}{\sum_{i} W_i} \right| \\
U &= \max(T - \max(F), 0) \\
L &= \max(\min(F) - T, 0) \\
SM &= \frac{1}{n} \sum_{i=1}^{n} (\mu_{\text{error}_i} + U_i + L_i)
\end{align}

where,
$W$ be the weights vector, $T$ be the target value, $F$ be the $i$th scenario, $\mu_{\text{error}}$ be the mean error calculated using weights, $U$ be the upper bound penalty, and $L$ be the lower bound penalty.

\begin{figure*}[htbp]
\centering
{\includegraphics[width=182mm]{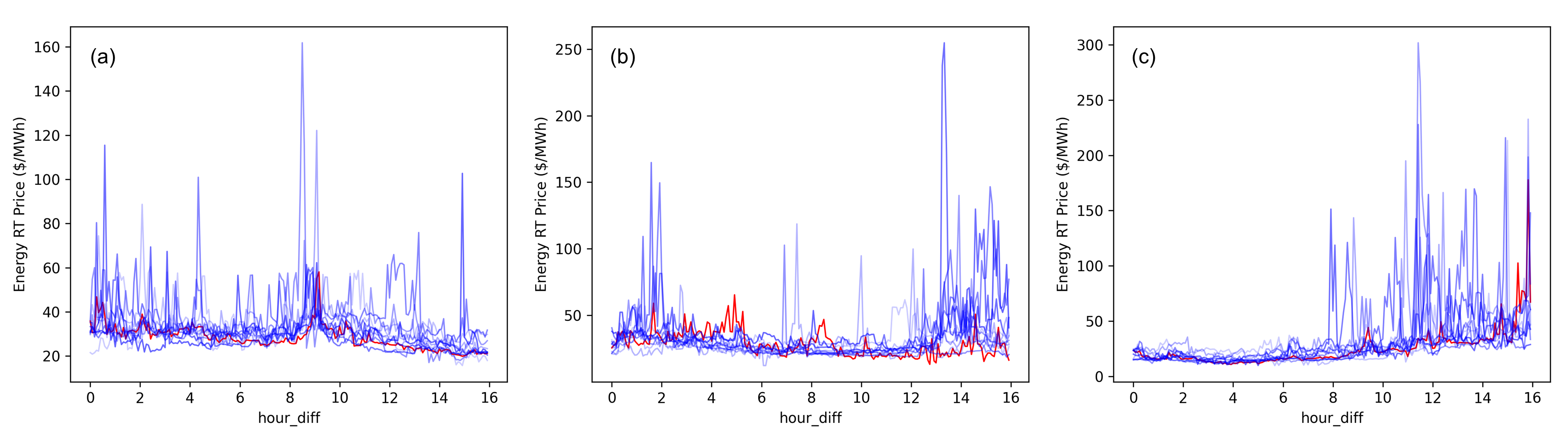}}
\caption{Scenarios of RT energy price from benchmark approach for representative days in PJM market: (a) 01/25/2023 09:00H (b) 04/14/2023 17:00H (c) 07/25/2023 00:00H }
\label{fig2}
\end{figure*}

Figure \ref{fig2} displays ten scenario subsets selected by the baseline method for the Energy RT product on representative days and hours: (a) 01/25/2023 09:00H (b) 04/14/2023 17:00H (c) 07/25/2023 00:00H. The true Energy RT values are represented by the red lines, while the various scenarios are depicted in shades of blue, with the intensity of the color indicating the probability of each scenario; darker hues denote higher probabilities, while lighter shades suggest lower probabilities. The abscissa delineates the optimization horizon. Within the chosen framework, the stochastic optimization agent forecasts over a 16-hour period to make an optimal decision for the subsequent hour.

From an analysis of Fig. \ref{fig2}, several observations emerge. Firstly, the scenarios tend to consistently over-predict, often reaching unnecessary extremes. Despite this trend towards over-prediction, the scenarios fail to accurately capture crucial price spikes, as exemplified in Fig. \ref{fig2}b. Moreover, the scenario depicted in Fig. \ref{fig2}c illustrates another issue: not only does it mispredict the magnitude, but it also inaccurately forecasts the timing of a significant spike. For a more quantitative assessment, the average $SM$ value for these scenarios was calculated to be 34.84. These discrepancies underscore the challenges in effectively modeling uncertainties and emphasize the need for more sophisticated prediction mechanisms.

\begin{figure*}[htbp]
\centering
{\includegraphics[width=182mm]{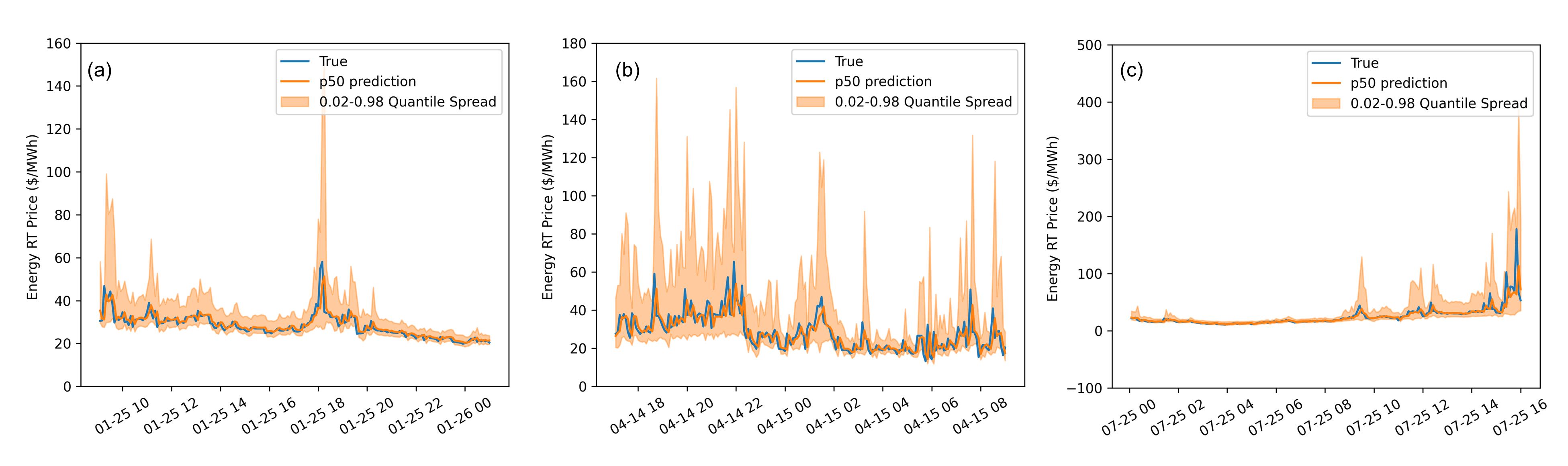}}
\caption{Energy RT forecast from transformer for (a) 01/25/2023 09:00H (b) 04/14/2023 17:00H (c) 07/25/2023 00:00H }
\label{fig3}
\end{figure*}

Figure \ref{fig3} showcases the probabilistic forecast of the Energy RT in the PJM market as generated by the Transformer model. The solid orange line delineates the median forecast, with the surrounding shade indicating the respective quantile range. Similar to previous illustrations, the forecasting horizon spans 16 hours, in line with the agent's optimization period for the subsequent day. Notably, the Transformer model exhibits commendable precision in predicting real-time Energy RT prices, with an average Mean Absolute Error (MAE) of approximately 3 \$/MWh. Leveraging such accurate forecasts could substantially enhance the performance of the stochastic model. However, a crucial aspect to consider is the multi-product nature of the optimization. Namely, different participation scenarios require anticipating not only energy price volatility for energy arbitrage opportunities, but also ancillary obligations and real-time ancillary prices, among others. Relying solely on independent forecasts for each product could disrupt the inherent coherence among them. This poses the intricate question: how can we integrate this forecast information to guide the historical subset selection more effectively? Our proposed approach seeks to answer this conundrum, and the subsequent paragraph delves into the demonstrated efficacy of scenarios built upon this method.

\begin{figure*}[htbp]
\centering
{\includegraphics[width=182mm]{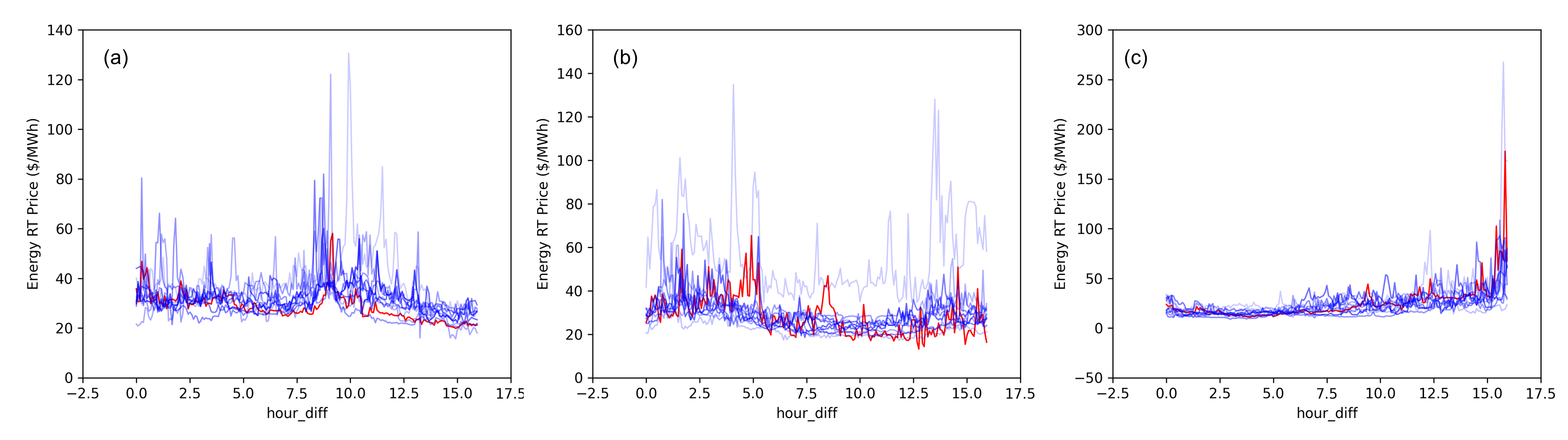}}
\caption{Scenarios of Energy RT product from the proposed approach for representative days in PJM market: (a) 01/25/2023 09:00H (b) 04/14/2023 17:00H (c) 07/25/2023 00:00H }
\label{fig4}
\end{figure*}

Figure \ref{fig4} illustrates the scenarios chosen using our proposed method, which incorporates both the Transformer model and wcDTW for selection. Evidently, the chosen scenarios closely envelop the true value, demonstrating improved precision. Moreover, the approach effectively addresses previously observed issues such as unwarranted over-predictions, accurate spike capture, and pinpointing the precise spike occurrence timings. From a quantitative perspective, the derived average $SM$ value stands at 27, marking an approximate 22\% enhancement in our preliminary assessment metric. This promising outcome paves the way for a deeper evaluation, conducted by running a stochastic simulation that determines battery bids in real-time. The tangible benefits of this improvement are further expounded in the subsequent section.

A more rigorous assessment of our method is undertaken by simulating the stochastic optimization process in intra-day bidding in actual market where the bidding prices are indeed influenced by the actions and strategies of other market participants. In this context, the agent determines whether to charge or discharge the stand-alone battery within the real-time market. This decision-making is conducted while adhering to the inherent constraints associated with the battery's physical limitations, market stipulations, as well as ensuring optimal battery life, managing state of charge (SoC), and complying with the real-time market's demand-supply balance. For the evaluation period spanning the month of July 2023 in the PJM market, the newly proposed technique yielded a revenue increase of 10\% compared to the preceding methodology. Table \ref{tab:my_label} delineates the revenue stratification based on charging or discharging activities for the Energy RT product. While other real-time products, such as ancillaries and regulations, contribute to the overall results, for the sake of brevity, we've exclusively spotlighted the Energy RT product. An examination of Table \ref{tab:my_label} reveals that our approach augments profitability by bolstering discharge-driven revenue and diminishing charge-associated costs. Assessment of $SM$ indicates a lasting impact and shows that July's improvement reflects an annual trend.


\begin{table}[htbp]
    \centering
        \caption{Revenue over July 2023 in real-time PJM market through stand-alone battery bidding}
    \begin{tabular}{|c|c|c|c|}
    \hline
        Product &  Dispatch &  Benchmark Revenue \cite{bhavsar2024stochastic} & Improved Revenue \\
         & & (\$/MWh)  & (\$/MWh) \\
        \hline
               Energy RT & Discharge & 26.67 & 30.97 \\
         & Charge  & -18.02 & -17.62\\
         \hline
    \end{tabular}

    \label{tab:my_label}
\end{table}

\section*{Conclusion}
The article presents a novel approach to intra-day battery bidding using a combination of Transformer-based forecasting and weighted constrained Dynamic Time Warping (wcDTW) for scenario selection. Our methodology leverages the predictive strength of Transformers, while wcDTW assists in effectively choosing representative scenarios from historical data. The rigorous simulations conducted in the PJM market during July 2023 affirmed the efficiency of our approach, which resulted in a revenue increase of 10\% compared to the benchmark method. Notably, our method improved profitability by maximizing discharge-driven revenue and minimizing charge-related costs. This paper substantiates the potential of integrating modern forecasting techniques with nuanced scenario selection mechanisms in enhancing the robustness and profitability of battery bidding in real-time energy markets.

\bibliographystyle{IEEEtran}
\bibliography{reference}
\end{document}